\documentclass{article}
\usepackage{multirow}
\usepackage{multicol}
\usepackage{hyperref}       
\usepackage{comment}
\usepackage{amsmath}
\usepackage{placeins}

  \usepackage[dblblindworkshop, final]{neurips_2025}
\workshoptitle{Machine Learning and the Physical Sciences}
\usepackage[pdftex]{graphicx}

\usepackage[utf8]{inputenc} 
\usepackage[T1]{fontenc}    
\usepackage{hyperref}       
\usepackage{url}            
\usepackage{booktabs}       
\usepackage{amsfonts}       
\usepackage{nicefrac}       
\usepackage{microtype}      
\usepackage{xcolor}         
\usepackage{wrapfig}

\title{PIANO: Physics-informed Dual Neural Operator for Precipitation Nowcasting}

\author{%
  Seokhyun Chin\\
  California Institute of Technology\\
  Pasadena, CA, 91125 \\
  \texttt{schin@caltech.edu} \\
  \And
   Junghwan Park\\
   TelePIX \\
    Seoul, South Korea, 07331  \\
  \texttt{brian897743@gmail.com} \\
   \And
  Woojin Cho\\
   TelePIX \\
Seoul, South Korea, 07331  \\
  \texttt{woojin.py@gmail.com} \\
}

\begin{document}

\maketitle

\begin{abstract}
Precipitation nowcasting, key for early warning of disasters, currently relies on computationally expensive and restrictive methods that limit access to many countries. To overcome this challenge, we propose precipitation nowcasting using satellite imagery with physics constraints for improved accuracy and physical consistency. We use a novel physics-informed dual neural operator (PIANO) structure to enforce the fundamental equation of advection-diffusion during training to predict satellite imagery using a PINN loss. Then, we use a generative model to convert satellite images to radar images, which are used for precipitation nowcasting. Compared to baseline models, our proposed model shows a notable improvement in moderate (4mm/h) precipitation event prediction alongside short-term heavy (8mm/h) precipitation event prediction. It also demonstrates low seasonal variability in predictions, indicating robustness for generalization. This study suggests the potential of the PIANO and serves as a good baseline for physics-informed precipitation nowcasting.

\end{abstract}

\section{Introduction}

As global warming intensifies, extreme precipitation events are becoming more common, leading to significant issues such as severe flooding, soil erosion, landslides, reduction of agricultural productivity, and increased health risks~\citep{Ombadi2023,Tabari2020}. To mitigate the impact of such problems, accurate precipitation nowcasting is essential. 

Traditional precipitation prediction relies heavily on Numerical Weather Prediction (NWP) models, but these suffer from heavy computational requirements and low spatial resolution, which hinders the prediction of heavily localized weather events. To combat this issue, many data-driven weather prediction methods have emerged. Models like Pangu-Weather~\citep{Bi2023}, FourCastNet~\citep{Kurth2023}, and GraphCast~\citep{Lam2023} have shown comparable performance to traditional NWP methods in data-driven approaches. However, these weather forecasting models rely heavily on computationally expensive ERA5~\citep{hersbach2020era5} reanalysis data for predictions, thus still require supercomputers for training. Many teams have  explored radar-image-based precipitation nowcasting methods~\citep{Andrychowicz2023,Ravuri2021,Zhang2023} as a remedy, but these models apply only to regions with radar coverage, limiting their applicability to developing countries and other areas that lack coverage. 

Precipitation nowcasting using satellite imagery emerges as a viable alternative, as satellite images have near-global coverage and are easily accessible. Indeed, a variety of studies (e.g.,~\citep{Lebedev2019,Park2025,Gruca2022}) have already explored this possibility. However, these studies fail to take into account the foundational physics equations of atmospheric dynamics: an idea that has proven to improve predictive capabilities of AI-based  weather prediction~\citep{Kochkov2024,verma2024}. To this end, we propose a proof-of-concept PIANO (Physics-Informed duAl Neural Operator) for precipitation nowcasting based on satellite imagery that aims to produce physically consistent precipitation forecasts while maintaining overall accuracy.

\section{Proposed Methods: Physics-informed Dual Neural Operator}

This section details the proposed PIANO, a novel architecture designed for satellite-based precipitation nowcasting. The core of PIANO lies in its unique dual structure, which decouples the task of data-driven temporal forecasting from the inference of latent physical dynamics. This separation allows for the explicit integration of physical principles, specifically the advection-diffusion equation, as a powerful inductive bias to regularize the learning process.

\begin{wrapfigure}{r}{0.50\textwidth}
\centering
    \vspace{-2em}
    \setlength{\tabcolsep}{5pt}
    \includegraphics[width=0.50\textwidth]{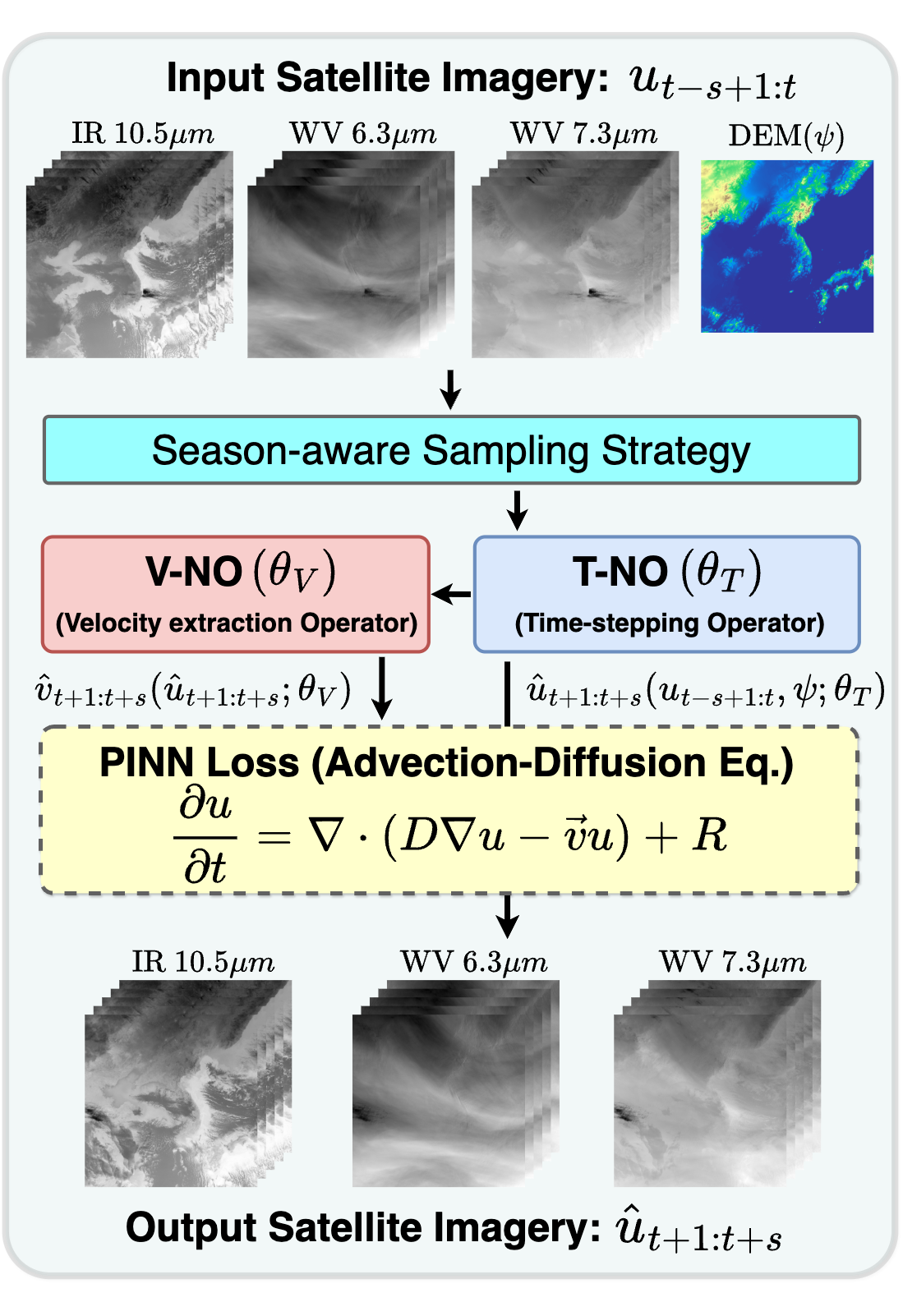}
    \caption{\textbf{Proposed method: PIANO.} The framework consists of two distinct neural operators: V-NO, which learns the mapping for velocity fields, and T-NO, which learns the mapping for time-stepping dynamics.}
    \vspace{-1em}
    \label{fig:model_pipeline}
\end{wrapfigure}
\subsection{Model Architecture}

As shown in the overall pipeline (Figure ~\ref{fig:model_pipeline}), the two operators run in series for the application of physics-informed neural networks (PINN) loss~\citep{raissi2019physics, dabrowski2023bayesian, cho2024parameterized} and data driven loss. The Time-stepping Neural Operator (T-NO) is a data-driven operator that learns to advance the state forward in time, while the Velocity-extraction Neural Operator (V-NO) is a physics-informed operator that learns the underlying flow field responsible for evolving the satellite imagery. By decoupling these tasks into separate modules, we can enforce physical consistency via V-NO without constraining the expressiveness of T-NO. 

T-NO takes input satellite images $u_{t-s+1:t}:\{u_{t-s+1},\cdots, u_{t-1}, u_t\}$ and $\psi$ where each $u\in\mathbb{R}^{C\times H \times W}$, represents a satellite image and $\psi\in\mathbb{R}^{H \times W}$ represents the digital elevation map. It outputs frames $\hat{u}_{t+1:t+s},$ which the V-NO takes as input to output $\hat{v}_{t:t+s}$. Both $\hat{u}_{t+1:t+s}$ and $\hat{v}_{t+1:t+s}$ are used to enforce a data-driven and PINN loss in a fashion similar to the physics-informed neural operator and the physics-informed deep neural operator~\citep{Goswami2023, Li2024PINO}. The loss function is described further in the section below.

We then use a Pix2Pix~\citep{isola2017image} ResNet~\citep{He2015} trained using a Generative Adversarial Network~\citep{goodfellow2014generative} structure with a ResNet generator and a ResNet-based discriminator network to transform predicted satellite images to radar images that can be used for precipitation nowcasting. The generative structure is motivated by the choice of~\citep{Park2025}, whose work we aim to improve with this study.
\subsection{Model Training}\label{sec:model_training}
We use a two-step training strategy where the V-NO and T-NO are pre-trained parallelly and combined for further fine tuning. In our study, T-NO takes eight time steps as input and output predictions of the next eight time steps, and V-NO takes eight time steps and outputs eight velocity fields.

\paragraph{Governing Equation: Advection-Diffusion Equation (Eq.~\eqref{eq:advection_diffusion})}  
Because we aim to model precipitation, we assume that the satellite images will evolve according to the governing advection-diffusion equation, which can be modeled as~\citep{Wesseling2001}: 
\begin{equation}
    \frac{\partial u}{\partial t}= \nabla \cdot (D\nabla u-\vec{v}u) +R,\label{eq:advection_diffusion}
\end{equation}
where ${\partial u}/{\partial t}$ represents the local rate of change of the quantity $u$ (in our case, the input of each layer in the satellite image), $\vec{v}$ is the velocity vector, $D$ is the diffusion coefficient and $R$ is the source term.
This equation can be discretized in time using finite differences, which allows us to achieve the following equation:
\begin{equation}
    u_{t+1} =  u_t +D \nabla^2 u_t - \vec{v}_t \cdot (\nabla u_t)-u_t (\nabla \cdot \vec{v_t}) +R.
\end{equation}
We utilize finite differencing due to the algorithmic efficiency and simplicity~\citep{Zhu2019}. In implementation, the values $u_{0:8}, D, \text{and } R$ are all scalar matrices of size $H \times W$, and the gradient and Laplacian of each value can be calculated using a convolution matrix~\citep{Woods2012}.  Under this implementation, the value $R$ represents the input of the quantity within each region and the diffusivity constant $D$ represents the diffusion between two regions. The beginning constants for the two values are 0 for $R$ and 1 for $D$, respectively.

We train V-NO to learn the velocity $\vec{\hat{v}}_{0:8}, D, R$ from the satellite images $u_{0:8}$ using the PINN loss only, which is written as:
\begin{equation}
    \mathcal{L}_{PDE}=\sum_t|| u_{t+1}- u_t -D \nabla^2 u_t +\\ \vec{\hat{v}}_t \cdot (\nabla u_t)+u_t (\nabla \cdot \vec{\hat{v}}_t) -R||^2
\end{equation}

The T-NO does not explicitly encode physics; it learns the spatio-temporal patterns of satellite imagery evolution purely from data. To train T-NO, we use a standard supervised reconstruction loss comparing the predicted frames to the ground-truth future frames. Let $\hat{u}_t$ denote the T-NO prediction for time $t$ and $u_t$ the corresponding ground truth. We minimize the mean squared error (MSE) over all predicted time steps in the training sequences, \[\mathcal{L}_{data} = \sum_t||\hat{u}_t-u_t||^2 \]

Finally, the V-NO and T-NO are combined and are trained with both the MSE and physics-informed loss functions, which can be written as:
\begin{equation}
    \mathcal{L}_{total}=  \sum_t||\hat{u}_t-u_t||^2 + \alpha ||\hat{u}_{t+1}- \hat{u}_t -D \nabla^2 \hat{u}_t + \vec{\hat{v}}_t \cdot (\nabla \hat{u}_t)+\hat{u}_t (\nabla \cdot \vec{\hat{v}}_t) -R||^2\label{eq:total_loss}
\end{equation}
\begin{equation}
    \mathcal{L}_{total}=  \mathcal{L}_{data}+\alpha \mathcal{L}_{PDE}
\end{equation}
\subsection{Dataset} 
We use the Sat2RDR~\citep{Park2025} dataset, which provides GK2A satellite infrared (10.5 µm), upper- (6.3 µm) and lower-level (7.3 µm) water vapor, plus DEM data of South Korea. We subsample January–August 2020–2024, excluding September–December due to missing 0600-hour values,  which makes training unbalanced temporally if incorporated correctly (i.e., times 0000-0500 would not be trained to a similar extent). For PIANO training, patches of size 768 × 768 are randomly cropped, with season-aware sampling~\citep{Park2025} yielding twenty frames per month. We reserve 2023 for validation and 2024 for testing. For the Pix2Pix ResNet, radar and satellite images are cropped to 224 × 224 and augmented with flips. Combined precipitation prediction performance is evaluated on January-August 2022 and 2023.

\subsection{Baseline models}
We select the two SOTA models relevant to satellite-image-based precipitation nowcasting, which are PhyDNet~\citep{Guen2020,Pihrt2022} and the Neural Prediction Model (NPM)~\citep{Park2025}. We do not select radar-to-radar (e.g., the model of~\citet{gao2023prediff}) or fusion-to-radar (e.g., the model of~\citet{Andrychowicz2023}) models due to it not performing our specific task. Models are trained on a singular RTX 5090 with a batch size of three.
\subsection{Evaluation Metrics}
In evaluating the precipitation prediction accuracy, we use the critical success index (CSI), as done by~\citet{Andrychowicz2023} and~\citet{Park2025}. The CSI is calculated as follows:
\begin{equation}
    CSI =\frac{TP}{TP+FP+FN},
\end{equation} where True Positives (TP) denote the number of correctly predicted precipitation events, False Positives (FP) denote incorrectly predicted precipitation events, and False Negatives (FN) denote the number of missed precipitation events. We specifically use the CSI of moderate (more than 4mm/h; CSI 4mm) and heavy (more than 8mm/h; CSI 8mm) precipitation as metrics as they are known to be more challenging.
\section{Results and Discussions}
In this section, we evaluate the validity of our proposed PIANO and compare the performance of it by comparing it against several baseline methods.

\begin{wrapfigure}{r}{0.7\textwidth}
    \centering
    \includegraphics[width=0.85\linewidth]{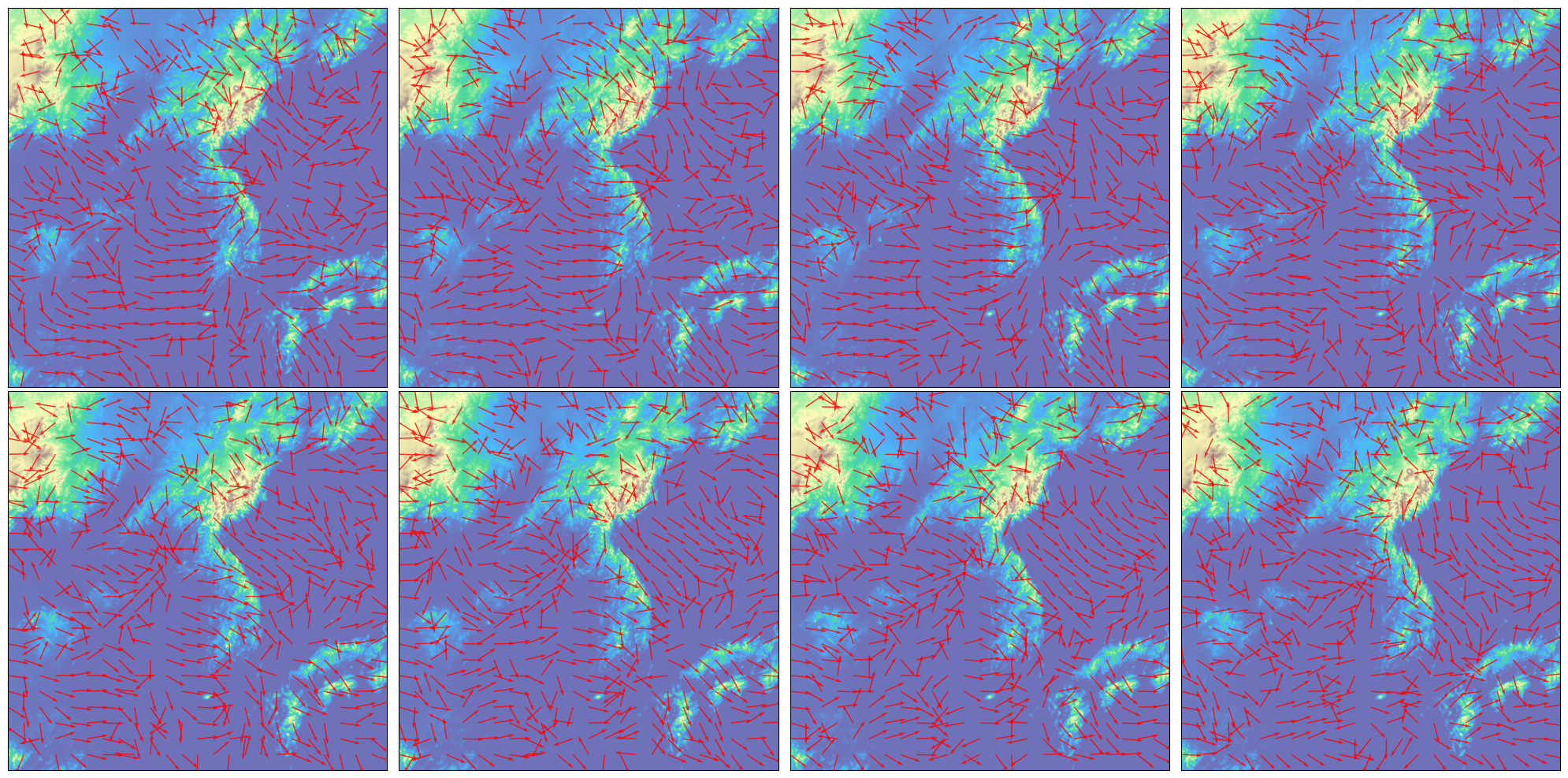}
    \caption{Visualization of the output of the V-NO on January 1st, 2024.}
    \label{fig:velocity_fields}
\end{wrapfigure}
\paragraph{V-NO outputs}
Figure~\ref{fig:velocity_fields} demonstrates the vector field outputted due to the V-NO. It can be noticed that all images have a generally rightward (note the bottom right parts of images) and downward trend (note the middle regions of images) in velocity, which is in line with the previously known trend of the atmosphere over South Korea~\citep{Baek2015}.

\paragraph{Precipitation Nowcasting Performance}
Table~\ref{tab:CSI_comparison} shows the CSI of each model when paired with the GAN. 
Our model is able to predict heavier precipitation events of 4mm with more skill than both baseline models, demonstrating our approach's validity and potential. Additionally, it performs well on predicting the first two hours of 8mm precipitation, although it becomes less accurate than PhyDNet in further prediction windows. This is a trend also observed in the work of~\citet{Park2025}, where PhyDNet outperforms the NPM in later time frames. It is to note that the CSI values calculated from our models are incredibly larger than those reported by~\citet{Andrychowicz2023} and~\citet{Park2025}. This may be due to the differently sampled data during training and evaluation of Sat2RDR performance that may have led to relatively simpler predictions. 

\begin{table}[!t]
    \centering
    \caption{Comparison of NPM, PhyDNet and our proposed method with CSI($\uparrow$).}
    \setlength{\tabcolsep}{3.3pt}
    \renewcommand{\arraystretch}{1.2}
    \resizebox{\textwidth}{!}{
    \begin{tabular}{ccccccccccccccccc}
    \specialrule{1pt}{2pt}{2pt}
    \multirow{2.5}{*}{Method} & \multicolumn{8}{c}{\textbf{CSI 4 mm}} & \multicolumn{8}{c}{\textbf{CSI 8 mm}} \\
    \cmidrule(lr){2-9}\cmidrule(lr){10-17}
        & 1h & 2h & 3h & 4h & 5h & 6h &7h & 8h& 1h & 2h & 3h & 4h & 5h & 6h & 7h & 8h \\
        \midrule
    NPM & 0.750 & 0.752 & 0.753 & 0.754 & 0.755 & 0.756 & 0.756 & 0.755
    & 0.599 & 0.599 & 0.599 & 0.600 & 0.600 & 0.600 & 0.600 & 0.599 \\
    PhyDNet & 0.737 & 0.743 & 0.744&0.747&0.752&0.756&0.760&0.763 &0.601 & 0.611 & 0.617 & 0.626 & 0.638 & 0.646 & 0.652 & 0.654  \\
    Ours & 0.757 & 0.758 & 0.760 & 0.757 & 0.760 & 0.763 & 0.763 & 0.763
     & 0.611 & 0.612 & 0.613 & 0.611 & 0.612 & 0.616 & 0.616 & 0.614 \\        
     \specialrule{1pt}{2pt}{2pt}
    \end{tabular}
    }
    \label{tab:CSI_comparison}
\end{table}
\newpage
\begin{wrapfigure}{l}{0.65\textwidth}
    \centering
\includegraphics[width=0.95\linewidth]{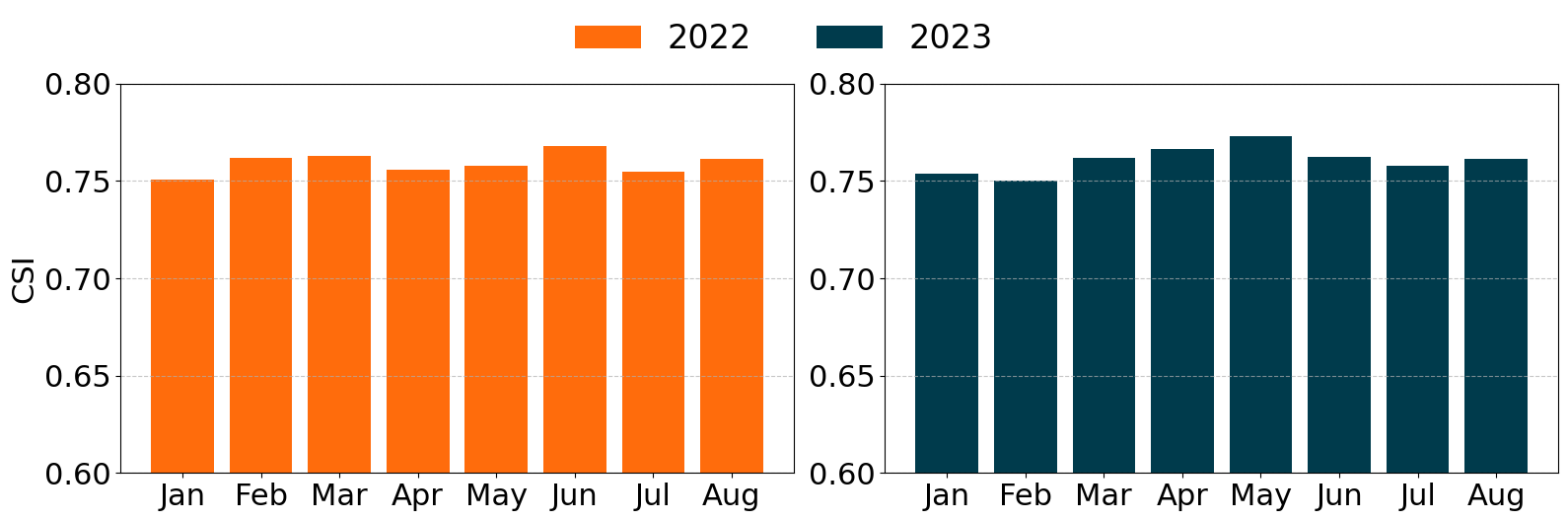}
    \caption{Performance comparison of CSI 4mm of PIANO by month.}
    \label{fig:permonth}
\end{wrapfigure}
\paragraph{Comparison of month}
Figure~\ref{fig:permonth} shows consistent CSI 4mm performance across all months, indicating strong precipitation nowcasting across the time domain. While there are some seasonal variations, the degree of change is low compared to the NPM, which showed incredibly high variation up to 0.1~\citep{Park2025}. 
\newline
\paragraph{Sensitivity Analysis}
Because the value of $\alpha$ in Eq.~\eqref{eq:total_loss} is important in determining training, we modify the value of $\alpha$ and train varying models to obtain results shown in Table~\ref{tab:ablation}. 
\begin{table}[!htb]
    \centering
    \small
    \caption{Satellite Prediction MSE of different iterations of PIANO}
    \begin{tabular}{cccccccccc}
        \specialrule{1pt}{2pt}{2pt}
        Model type& 1h & 2h & 3h & 4h & 5h & 6h & 7h & 8h\\\midrule
        PIANO ($\alpha=5.0$) & 0.247 & 0.269 &  0.316 &  0.369 &  0.424 & 0.480 &  0.535 & 0.592\\PIANO ($\alpha=1.0$) & 0.205 & \textbf{0.252} & \textbf{0.309} & \textbf{0.366} & \textbf{0.419} & \textbf{0.471} & \textbf{0.520} & \textbf{0.569}\\ 
        PIANO ($\alpha=0.2$) & 0.205 & 0.261 & 0.321 & 0.376 & 0.431 & 0.482 & 0.532 & 0.580  \\
        PIANO ($\alpha=0.0$) & \textbf{0.202} & 0.260 & 0.320 & 0.375 & 0.430 & 0.480 & 0.529 & 0.576  \\
        \specialrule{1pt}{2pt}{2pt}
    \end{tabular}
    \label{tab:ablation}
\end{table}
\\
It is clear that the value of $\alpha$ must be selected carefully; both over ($\alpha=5$) and under ($\alpha=0.2$) constraining the T-NO produces poor results. However, if $\alpha$ is selected well, using the V-NO and adding PINN loss improves performance of the model in predicting further lead times. However, it is to note that the T-NO produces highest performance at a lead time of one hour. This is likely because the PINN loss does not constrain the prediction value of the first hour at all, therefore the PINN loss does not help improve performance. 

\section{Conclusion}
We presented PIANO, a physics-informed dual neural operator framework for satellite-based precipitation nowcasting that cleanly separates two roles: V-NO estimates the velocity field over satellite imagery, while T-NO advances the state forward in time. Empirically, PIANO demonstrates competitive accuracy and seasonal consistency while remaining simple to train and deploy on satellite imagery, demonstrating its potential for future development. Focus on further optimization of model hyperparameters and comprehensive evaluations against strong baselines will be key to fully realizing PIANO’s capabilities.

{
\small

\bibliographystyle{unsrtnat}
\bibliography{citations}
}


\appendix
\setcounter{figure}{0}
\renewcommand{\thefigure}{A.\arabic{figure}}
\newpage
\section{Results in satellite to satellite image prediction}
\begin{figure}[!htb]
    \centering
    \includegraphics[width=\linewidth]{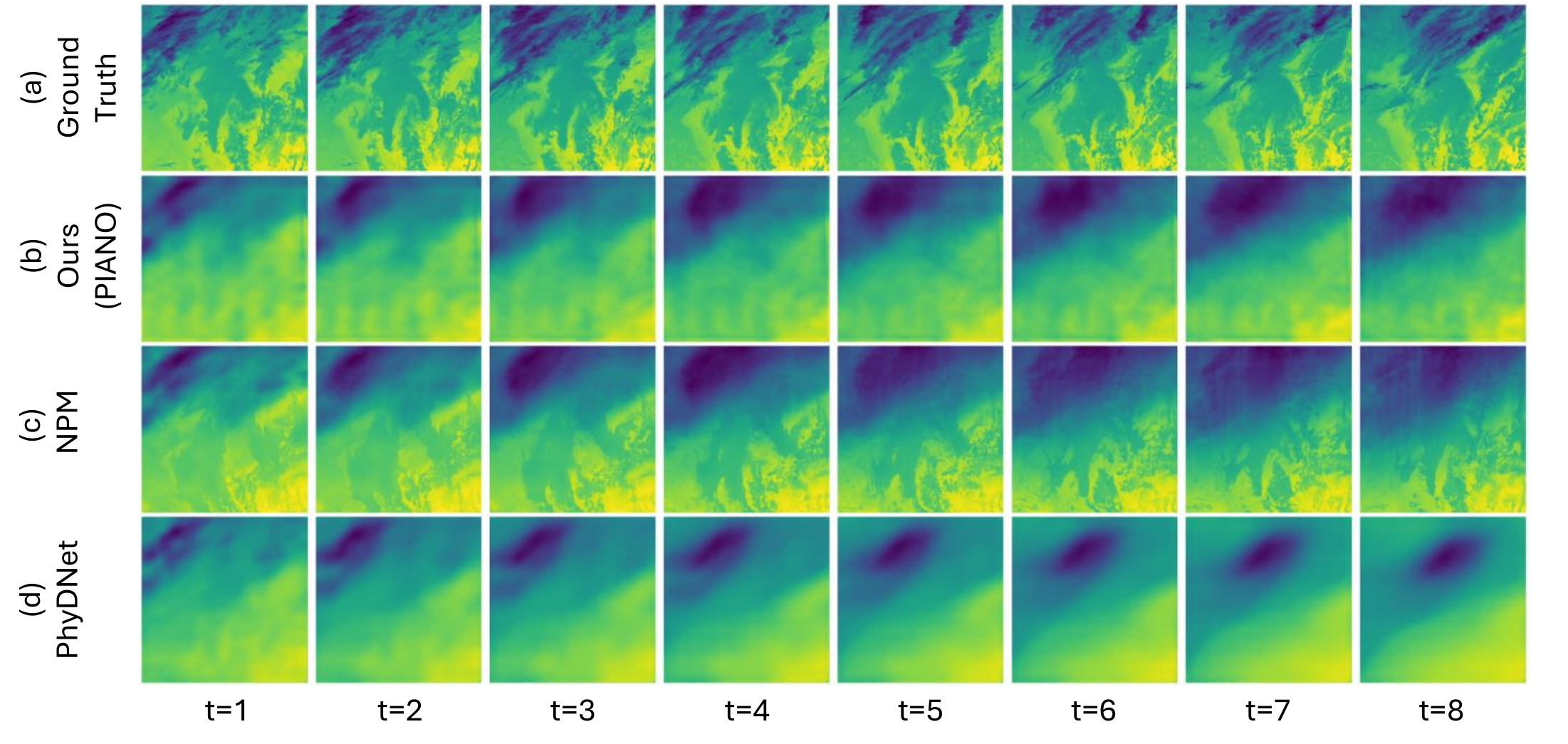}
    \caption{Visualization of the ground truth satellite IR band (a) and the prediction by the PIANO (b), NPM (c) and PhyDNet (d) for January 1st, 2024, 0900-1600 hrs}
    \label{fig:Appendix1}
\end{figure}
\begin{figure}[!htb]
    \centering
    \includegraphics[width=\linewidth]{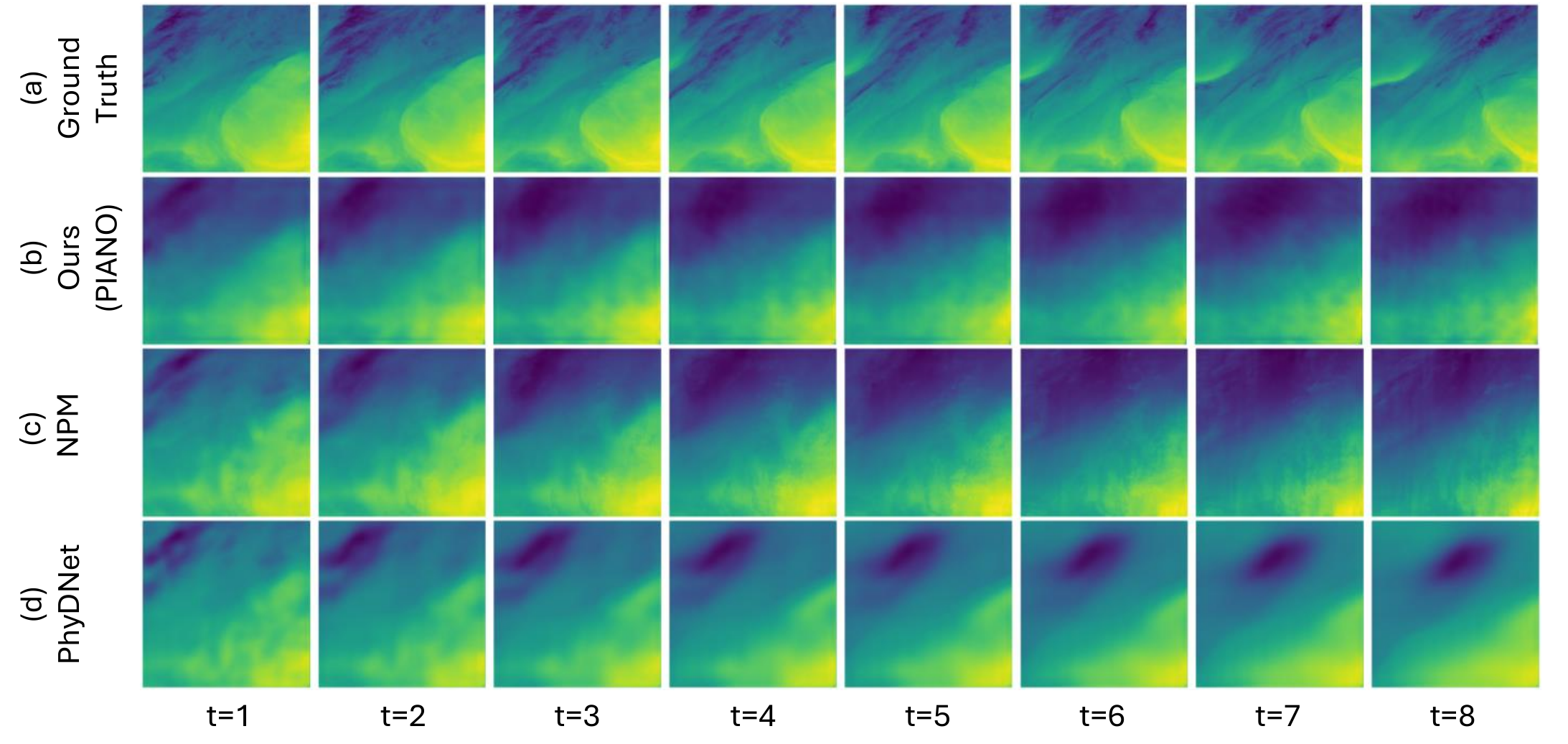}
    \caption{Visualization of satellite WV at 6.3 $\mu m$ band (a) and the prediction by the PIANO (b), NPM (c) and PhyDNet (d) for January 1st, 2024, 0900-1600 hrs}
    \label{fig:Appendix2}
\end{figure}
\begin{figure}[!htb]
    \centering
    \includegraphics[width=\linewidth]{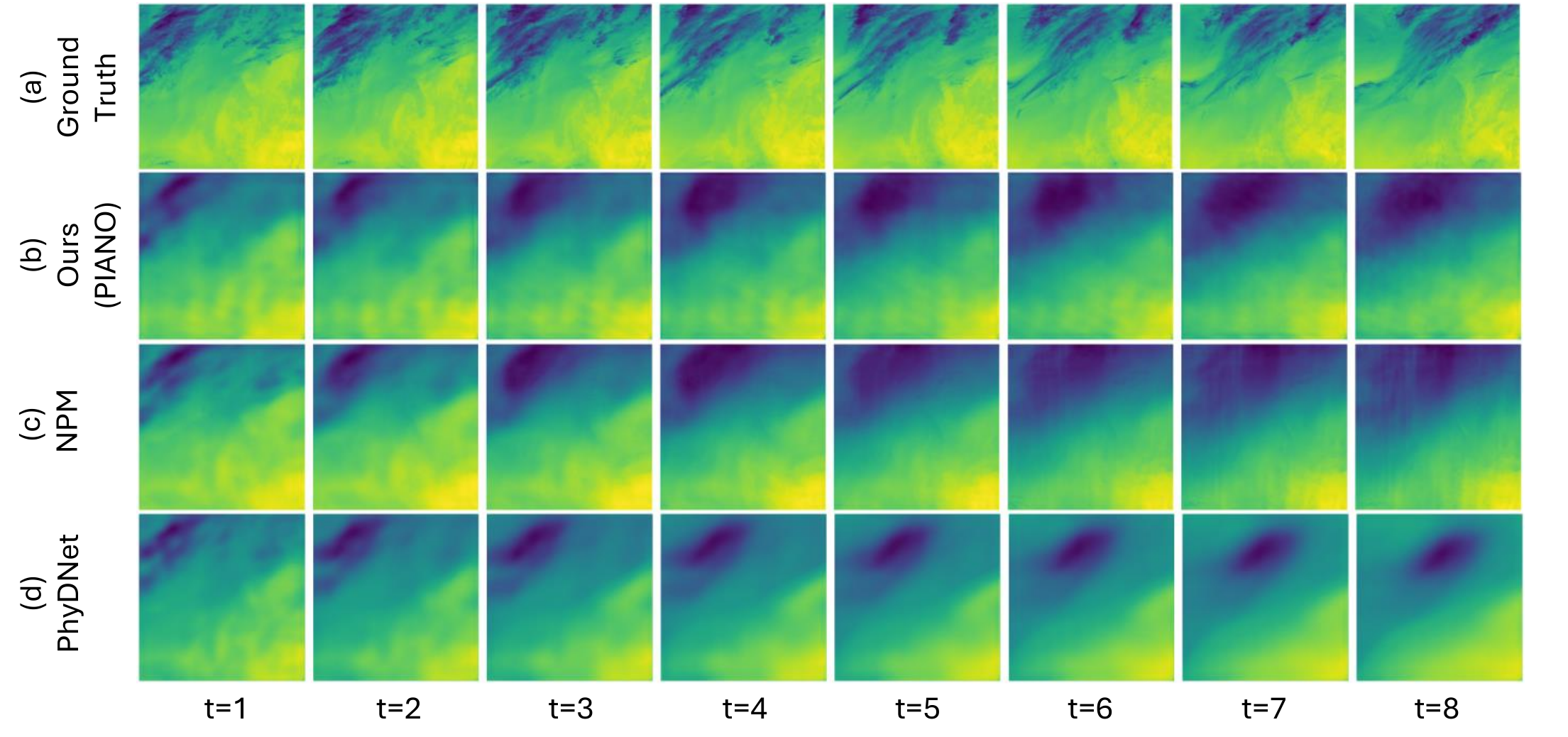}
    \caption{Visualization of satellite WV at 7.3 $\mu m$ band (a) and the prediction by the PIANO (b), NPM (c) and PhyDNet (d) for January 1st, 2024, 0900-1600 hrs}
    \label{fig:Appendix3}
\end{figure}
\FloatBarrier
\section{Results in satellite to radar transformation}
\setcounter{figure}{0}
\renewcommand{\thefigure}{B.\arabic{figure}}
\begin{figure}[!htb]
    \centering
    \includegraphics[width=\linewidth]{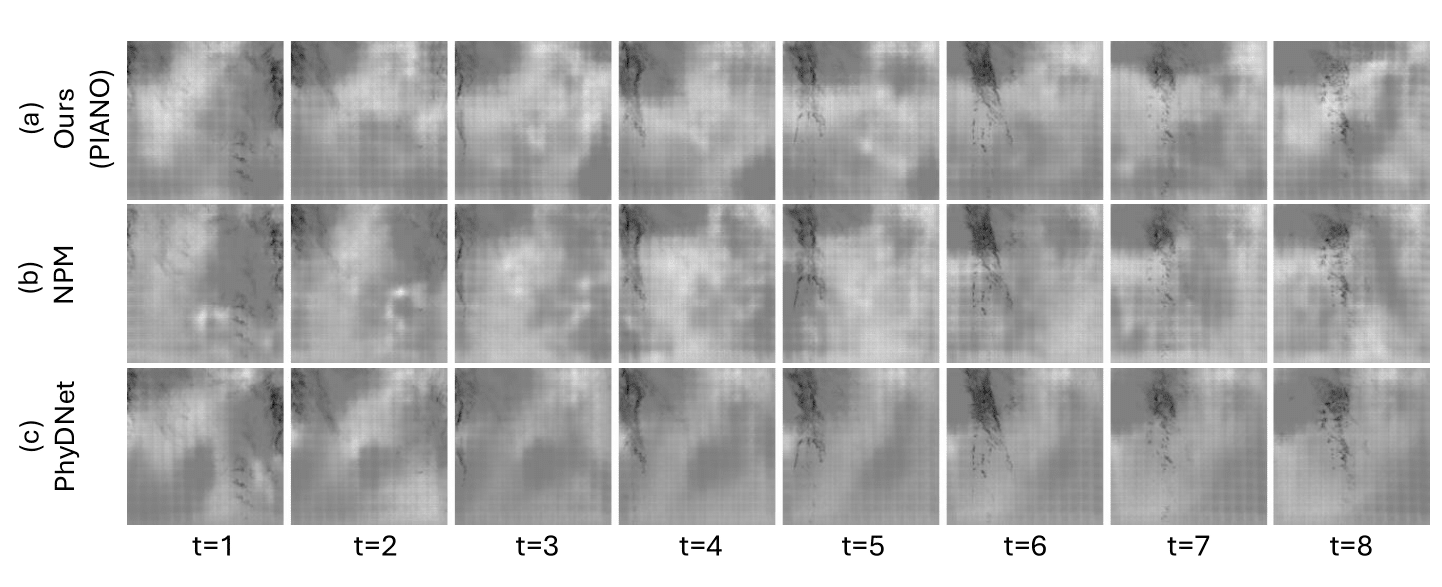}
    \caption{Visualization of difference between ground truth radar and PIANO predictions(a), NPM predictions (b) and PhyDNet predictions(c) for April 23rd, 2024, 1400-2100 hrs. The date and time was selected due to the recorded high precipitation within that window}
    \label{fig:Appendix4}
\end{figure}
\end{document}